\pdfoutput=1

\documentclass[11pt]{article}

\usepackage{acl}
\usepackage{times}
\usepackage{latexsym}
\usepackage[T1]{fontenc}
\usepackage[utf8]{inputenc}
\usepackage{microtype}
\usepackage{booktabs}
\usepackage{graphicx}
\usepackage{multirow}
\usepackage{hyperref}
\usepackage{amsmath,amssymb}
\title{
Generate, Discriminate and Contrast:\\
A Semi-Supervised Sentence Representation Learning Framework
}

\author{Yiming Chen$^{\dag}$ \quad Yan Zhang$^{\dag}$ \quad Bin Wang$^{\dag}$ \quad Zuozhu Liu$^{\ddag  \star}$\thanks{\quad Corresponding author.} \quad Haizhou Li$^{\natural, \dag, \S}$ \\
        $^\dag$National University of Singapore \quad
        $^{\ddag}$Zhejiang University $^{\star}$Angelalign Inc., China\\
        $^{\natural}$The Chinese University of Hong Kong, Shenzhen, China \quad $^{\S}$Kriston AI Lab, China\\
        \tt yiming.chen@u.nus.edu,  {zuozhuliu}@intl.zju.edu.cn\\
        \tt \{haizhou.li,eleyanz,bin.wang\}@nus.edu.sg \\
}

\begin{document}

\maketitle






\begin{abstract}

Most sentence embedding techniques heavily rely on expensive human-annotated sentence pairs as the supervised signals. Despite the use of large-scale unlabeled data, the performance of unsupervised methods typically lags far behind that of the supervised counterparts in most downstream tasks. In this work, we propose a semi-supervised sentence embedding framework, GenSE, that effectively leverages large-scale unlabeled data. Our method include three parts: 1) Generate: A generator/discriminator model is jointly trained to synthesize sentence pairs from open-domain unlabeled corpus; 2) Discriminate: Noisy sentence pairs are filtered out by the discriminator to acquire high-quality positive and negative sentence pairs; 3) Contrast: A prompt-based contrastive approach is presented for sentence representation learning with both annotated and synthesized data. Comprehensive experiments show that GenSE achieves an average correlation score of 85.19 on the  STS datasets and consistent performance improvement on four domain adaptation tasks, significantly surpassing the state-of-the-art methods and convincingly corroborating its effectiveness and generalization ability.\footnote{Code, Synthetic data and Models available at \url{https://github.com/MatthewCYM/GenSE}}  

\end{abstract}     
\section{Introduction}

Sentence representation learning has recently attracted outspread research attention. It learns vector representations for sentences, which can be subsequently utilized on a wide range of downstream tasks, including information retrieval~\citep{thakur2021beir,justrank-2022,misra-etal-2016-measuring}, language understanding and evaluation~\citep{cer2018universal,conneau-kiela-2018-senteval,perone2018evaluation,zhang-etal-2021-dynaeval}. For sentence representation learning, the contrastive learning-based approaches, including supervised and unsupervised ones, have demonstrated to be the most efficient and effective~\citep{zhang-2020-unsupervised,carlsson2021semantic, giorgi-etal-2021-declutr,gao-etal-2021-simcse}. In contrastive learning, the quality of positive and negative pairs has a large impact on the overall performance~\citep{chen2020simple,gao-etal-2021-simcse}. In particular, previous supervised contrastive methods usually construct sentence pairs using human-annotated natural language inference (NLI) data~\citep{bowman-etal-2015-large, williams-etal-2018-broad}, and outperform unsupervised approaches by a large margin~\citep{gao-etal-2021-simcse}. However, the state-of-the-art supervised methods usually rely on multi-source labeled data to generalize to various downstream tasks~\citep{reimers-gurevych-2019-sentence}, while such large-scale annotated data across domains are not always available.

Recent works also attempt to leverage the resources of unlabeled sentences for better sentence representation learning. A straightforward choice is to adopt unsupervised or self-supervised paradigms, such as BERT-flow~\citep{li-etal-2020-sentence}, SimCSE~\citep{gao-etal-2021-simcse}. However, the performance of these methods is still far behind the supervised counterparts. Another stream of work employs retrieval strategies to obtain close sentences as potential entailment pairs, and then trains a discriminator to re-label the retrieved sentence pairs ~\citep{thakur-etal-2021-augmented}. They attain satisfactory results but with two main limitations:  1) The retrieved sentences may not be the ideal entailment to the input sentences due to the limited numbers of sentences in the banking corpus, leading to low-quality synthetic sentence pairs; 2) The retrieval-based methods can only obtain entailment pairs but not contradiction ones, which can be an important source of hard negative samples.

\begin{figure*}[ht]
    \centering
    \includegraphics[scale=0.6]{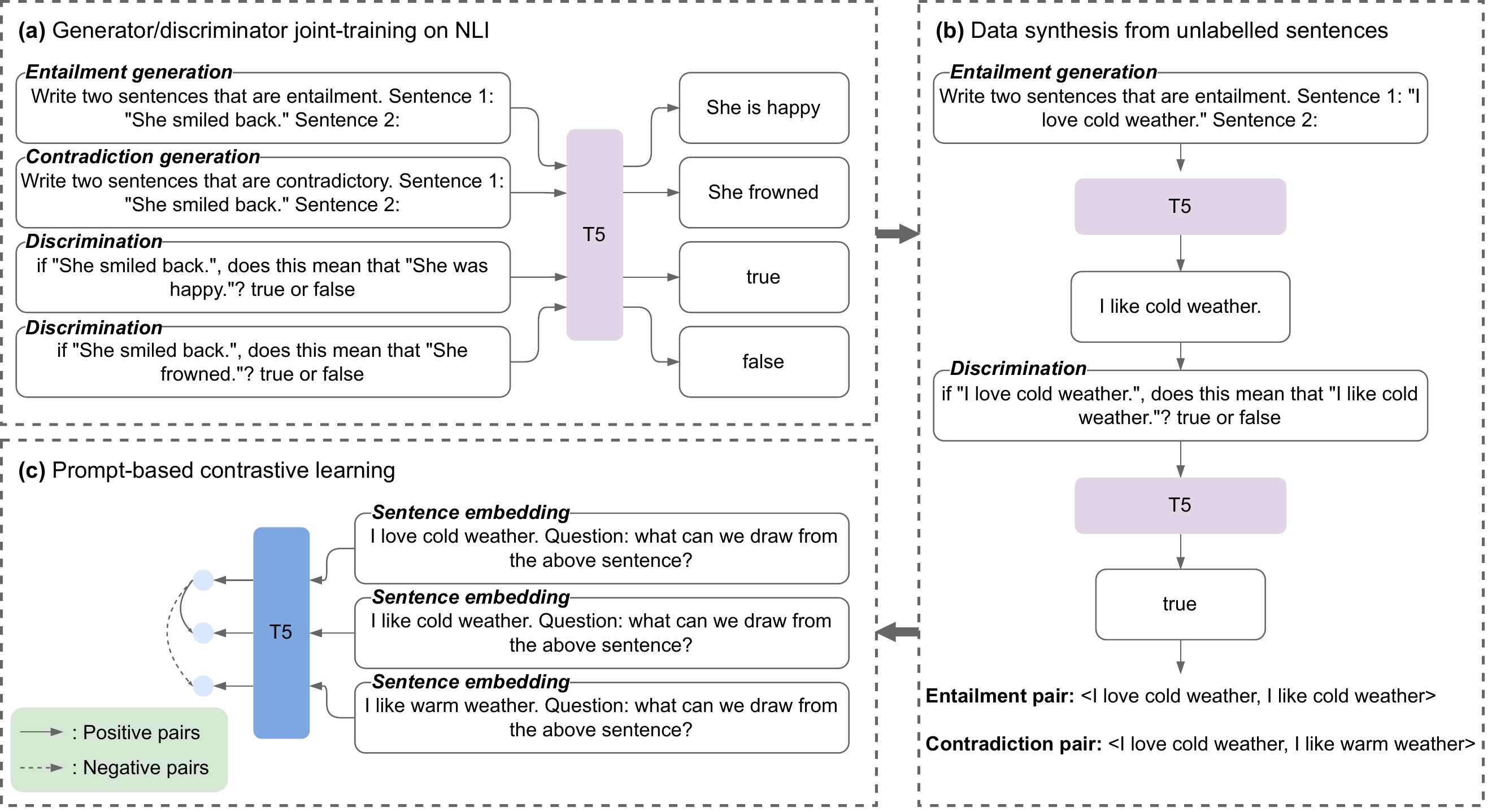}
    \vspace{-1mm}
    \caption{GenSE framework includes three steps: jointly training of generator/discriminator; data synthesis from unlabelled data; prompted-based contrastive learning. All models are initialized from same pre-trained weights. The model with the same colour shares the same weights. The upper-left of each sentence block refers to the prompt name listed in Table~\ref{tab:gd_templates}.} 
    \vspace{-3mm}
    \label{fig:all_framework}
\end{figure*}

In this paper, we propose a novel semi-supervised sentence representation learning framework leveraging both annotated and unlabeled corpus to address the aforementioned issues. Our method, GenSE, is built upon pre-trained text-to-text models. It integrates three tasks, i.e., generation, discrimination, and contrastive learning, into a single framework. Specifically, we first train a unified generator/discriminator model from NLI data, which is responsible for sentence pair generation and noisy pair discrimination. Afterwards, the open-domain unlabeled sentences are taken as inputs to the generator to synthesize sentence pairs, which are further discriminated to obtain high-quality positive and negative pairs. Finally, a prompt-based encoder-decoder model is trained in a contrastive manner to learn sentence embeddings from both human-annotated and our synthesized sentence pairs. 

We evaluate GenSE on the standard semantic textual similarity (STS) benchmark and four downstream domain adaptation tasks. On the STS benchmark, GenSE achieves an averaged Spearman's correlation of 84.78, which is further boosted to 85.19 by integrating additional QA data, significantly outperforming the state-of-the-art baselines. On domain adaptation tasks, we consider two different settings, i.e. direct transfer and domain adaptation. Our GenSE achieves an average 1.9\% improvement over baselines for direct transfer. Through further domain adaptation with sentence pairs synthesized from unlabeled in-domain data, GenSE achieves an extra 1.7\% improvement, confirming its adaptability for various downstream tasks.
Comprehensive ablation studies on different components of GenSE, extensive comparisons between data synthesis strategies, and in-depth analysis on the uniformity/alignment and quality of the synthetic data, convincingly validate the effectiveness and generalization ability of our GenSE framework. 
\begin{table*}
\centering

\resizebox{\textwidth}{!}{
\begin{tabular}{lll}
\toprule
\textbf{Tasks}           & \textbf{Input templates}                                                       & \textbf{Output templates} \\
\midrule
$T_{e}$: Entailment generation    & Write two sentences that are entailment. Sentence 1: "\textbf{{[}X1{]}}" Sentence 2:    & \textbf{{[}X2{]}}                  \\
$T_{c}$: Contradiction generation & Write two sentences that are contradictory. Sentence 1: "\textbf{{[}X1{]}}" Sentence 2: & \textbf{{[}X2{]} }                 \\
$T_{d}$: Discrimination           & if "\textbf{{[}X1{]}}", does this mean that "\textbf{{[}X2{]}}"? true or false                   & true/false \\
$T_{s}$: Sentence embedding      & \textbf{[X]} Question: what can we draw from the above sentence? & sentence embedding vector \\
\bottomrule
\end{tabular}
}

\caption{Prompt used in GenSE. [X] refers to the placeholder for input/output sentences.}
\label{tab:gd_templates}
\end{table*}

\section{Methodology}
The GenSE comprises two neural models, including a unified generator/discriminator model for sentence pair generation and quality check, and a sentence embedding model optimized with a contrastive objective. The learning schema of these two models consists of three consecutive steps, as illustrated in Figure~\ref{fig:all_framework}.

Firstly, we perform joint-training of the generator/discriminator model based on NLI dataset. 
Afterwards, we take the large-scale unlabeled sentences as inputs to generate entailment and contradiction pairs with the trained generator, which are further filtered by the discriminator to keep high-quality positive and negative pairs. Finally, we train the contrastive learning model based on the synthesized and human-annotated sentence pairs. We formulate all tasks into a text-to-text format with a prompt as the task signal. Therefore, we can build our models from the initialization of the same pre-trained text-to-text model, which leads to high modeling simplicity. Table~\ref{tab:gd_templates} lists all the input, output templates used in GenSE, including entailment generation template $T_{e}$, contradiction generation template $T_{c}$, discrimination template $T_{d}$, and sentence embedding template $T_{s}$. Below we'd like to elaborate more.

\subsection{Generate and Discriminate}
\label{method:gd}
The training process of the generator/discriminator model is illustrated in Figure~\ref{fig:all_framework}.(a). The model is trained to perform two tasks, i.e., generation and discrimination, simultaneously with NLI data. For the generation task, we first obtain two training instances for each NLI triplet $\{x_{ori},x_{entail},x_{contra}\}$. In particular, as for entailment, we place $x_{ori}$ into the pre-defined entailment input templates $T_{e}$ to obtain model input $T_e(x_{ori})$, and set the output label as the corresponding entailment hypothesis $x_{entail}$. Similarly, we use $T_{c}$ in the contradiction generation to obtain the training instance: $\{T_c(x_{ori}), x_{contra}\}$. The generator is trained to predict the output labels given the input prompts and sentences, as shown in the top of Figure~\ref{fig:all_framework}.(a). 



For the discrimination task, we apply the prompt template $T_{d}$ on the concatenated sentence pairs to obtain model inputs, and the output is either true or false, leading to two training instances: $\{T_d(x_{ori}, x_{entail}), \text{true}\}$ and $\{T_d(x_{ori}, x_{contra}), \text{false}\}$. The output is mapped from the annotated labels, i.e., entailment $\rightarrow$ true, and contradiction $\rightarrow$ false. 

By adding the prompts, both the generation and discrimination tasks can be transformed into conditional generation tasks, which largely reduce the model complexity. The model can be trained with a standard conditional generation loss to maximize the probability of the output sequence $y_{1,...,M}$ given the input sequence $X$:
\begin{equation}
\begin{split}
    P(y_1, & y_2,...,y_M|X) = \\ &\prod_{m=1}^{M}P(y_m|y_0,...,y_{m-1},X),
\end{split}
\label{eq:teacher_forcing}
\end{equation}
where $y_0$ is the decoder start token.

To better balance the performance between the generation and discrimination tasks, we equally mix the generation and discrimination training instances, and employ a weighted sum of the generation perplexity and discrimination accuracy on the development set for model selection. 

The data synthesis and quality check process is shown in Figure~\ref{fig:all_framework}.(b). Given an unlabelled sentence $u$, we first augment it with the generation prompts to obtain model inputs $T_e(u)$ and $T_c(u)$, which are fed into the trained generator to get entailment/contradiction prediction $u_{entail}$ and $u_{contra}$. The generated triplets $\{T_d(u_{ori}, u_{entail}),\text{true}\}$ and $\{T_d(u_{ori}, u_{contra}),\text{false}\}$ are evaluated by the discriminator for quality check, where the output probabilities of 'true' and 'false' are compared with a predefined threshold $\alpha$ to filter out the low-quality triplets.  Finally, we can get a clean synthetic corpus with both positive and negative pairs for sentence representation learning.



\subsection{Prompt-based Contrastive Learning}
Contrastive learning has been widely used in sentence representation learning, achieving state-of-the-art performance on the STS benchmark~\citep{gao-etal-2021-simcse, abs-2108-08877}. Recently, PromptBERT~\citep{abs-2201-04337} utilizes prompts to obtain sentence embeddings of much higher quality from encoder-based models. Inspired by it, we propose an improved prompt-based contrastive sentence representation model for text-to-text models. 

Our prompt-based contrastive learning is shown in Figure~\ref{fig:all_framework}.(c). we first augment the input sentence [X] with prompt $T_{s}$ shown in Table~\ref{tab:gd_templates}, and feed the "[PAD]" token into the decoder. We take the first decoder output as the sentence embedding as in~\citep{abs-2108-08877}. During training, we regard the entailment pairs as positive pairs, and the contradictory sentences as hard negative pairs. We also use in-batch negative sampling to include more negative samples during training. For a batch with $N$ triplets, the prompt-based contrastive learning loss function is defined as:


\begin{equation} 
    \mathcal{L} = \frac{e^{{\rm sim}(h_i,h_i^+)/\tau}}{\sum_{j=1}^{N}(e^{{\rm sim}(h_i,h_i^+)/\tau}+e^{{\rm sim}(h_i,h_i^-)/\tau})},
    \label{eq:loss}
\end{equation}
where $i,j$ is the sentence index, $\tau$ is the temperature hyper-parameter, ${\rm sim}(\cdot)$ is the cosine similarity function, and $\{h,h^+,h^-\}$ are representations of the premise, and  corresponding entailment and contradiction hypotheses. 

\subsection{Training Settings}
We consider three different settings, including universal sentence embedding, domain adaptation and QA training. 

\textbf{Universal sentence embedding:} We consider the general case where we have large-scale open domain unlabeled sentences and restricted amount of human-labeled NLI data. In this case, the GenSE is trained with a two-stage schema. In particular, we first train the sentence embedding model on the synthetic open-domain triplets from the generator/discriminator model, which is subsequently finetuned on the labeled NLI datasets.

\textbf{Domain adaptation:} The GenSE can also be used for domain adaptation. In this case, we assume the GenSE would have access to extra in-domain unlabeled data from the downstream tasks, which is regarded as a standard setting in unsupervised sentence embedding~\citep{wang-etal-2021-tsdae-using}. The open-domain corpus are available as well. We follow the same procedure in Section~\ref{method:gd} to obtain in-domain sentence triplets by taking the in-domain sentences as the input to the trained generator/discriminator model. Afterwards, the prompt-based sentence embedding model is trained in three stages: 1) Pretraining on open-domain synthetic pairs; 2) Adapting the sentence embedding to a specific domain using in-domain synthetic pairs; 3) Finetuning on labeled NLI data.

\textbf{QA training:} It is also not uncommon to use additional QA data to improve sentence representation learning. Here, we also develop a setting to demonstrate whether GenSE can further boost the performance along with labeled QA data. Since hard negatives are not available in QA data, we remove them from the loss function:
\begin{equation}
    \mathcal{L}' = \frac{e^{{\rm sim}(h_i,h_i^+)/\tau}}{\sum_{j=1}^{N}e^{{\rm sim}(h_i,h_i^+)/\tau}}.
    \label{eq:loss_wo_negative}
\end{equation}
We consider two ways to train the prompt-based contrastive learning model to demonstrate the effectiveness of GenSE: 1) GenSE-QA: the model is first trained on QA data and then finetuned on NLI data. 2) GenSE+: the model is first pretrained on the open-domain synthetic data generated by the generator/discriminator model, and then trained on the QA data, and finally finetuned on the NLI data. 

\begin{table*}
\centering
\small

\begin{tabular}{lccccccccc}
\toprule
\textbf{Model} & \textbf{\# Params} & \textbf{STS12} & \textbf{STS13} & \textbf{STS14} & \textbf{STS15} & \textbf{STS16} & \textbf{STSb} & \textbf{SICK-R} & \textbf{Avg} \\
\midrule
\multicolumn{10}{c}{\textit{Large-scale sentence embedding models}} \\
\midrule
SimCSE-RoBERTa-Large & 354M  & 77.46 & 87.27 & 82.36 & 86.66 & 83.93 & 86.70 & 81.95 & 83.76 \\
ST5-Enc-Large        & 335M  & 76.52 & 85.75 & 81.01 & 87.13 & 83.26 & 85.45 & 79.85 & 82.71 \\
ST5-EncDec-Large     & 335M  & 79.15 & 87.42 & 83.61 & 87.64 & 83.92 & 86.35 & 80.64 & 84.11 \\
ST5-Enc-Large-CommQA & 770M  & 79.10 & 87.32 & 83.17 & 88.27 & 84.36 & 86.73 & 79.84 & 84.11 \\
ST5-EncDec-3B        & 3B    & 79.24 & 87.80 & 83.95 & 87.75 & 84.60 & 86.62 & 80.91 & 84.41 \\
ST5-Enc-3B-CommQA    & 1.24B    & 79.02 & 88.80 & 84.33 & 88.89 & 85.31 & 86.25 & 79.51 & 84.59 \\
\midrule
\multicolumn{10}{c}{\textit{Base-size sentence embedding models}} \\
\midrule
SBERT-Base          & 110M & 70.97 & 76.53 & 73.19 & 79.09 & 74.30 & 77.03 & 72.91 & 74.89 \\
SimCSE-RoBERTa-Base & 110M & 76.53 & 85.21 & 80.95 & 86.03 & 82.57 & 85.83 & \textbf{80.50} & 82.52 \\ 
ST5-Enc-Base        & 110M & 77.37 & 83.65 & 80.41 & 86.04 & 81.70 & 84.49 & 79.79 & 81.92 \\
ST5-EncDec-Base     & 220M & 77.90 & 85.62 & 82.24 & 86.81 & 82.13 & 84.98 & 79.97 & 82.81 \\
ST5-Enc-Base-CommQA & 110M & 78.05 & 85.84 & 82.19 & 87.46 & 84.03 & 86.04 & 79.75 & 83.34 \\
\midrule
GenSE               & 220M & 80.72 & 87.43 & 83.96 & 88.63 & 85.19 & 87.65 & 79.87 & 84.78 \\
GenSE-QA            & 220M & \textbf{80.84} & 87.52 & 83.19 & 87.48 & 84.35 & 86.42 & 79.73 & 84.22 \\
GenSE+              & 220M & 80.65 & \textbf{88.18} & \textbf{84.69} & \textbf{89.03} & \textbf{85.82} & \textbf{87.88} & 80.10 & \textbf{85.19} \\
\bottomrule 
\end{tabular}

\caption{Results of sentence embedding on STS tasks. Spearman's correlation is reported. The first block shows previous state-of-the-art large-scale embedding models, and the second block shows results from base-size models.}
\label{tab:main_results}
\end{table*}


\section{Experiments}
\subsection{Experiment Setup}
We evaluate GenSE on seven popular STS tasks for universal sentence embedding, and on four datasets from various domains, including AskUbuntu~\citep{lei-etal-2016-semi}, CQADupStack~\citep{hoogeveen2015cqadupstack}, Twitter~\citep{xu-etal-2015-semeval, lan-etal-2017-continuously}, and BIOSSES~\citep{10.1093/bioinformatics/btx238}. For all the experiments, we assume the only available labeled sentence pairs are NLI (MNLI+SNLI)~\citep{bowman-etal-2015-large, williams-etal-2018-broad}. For open-domain data synthesis, we sample sentences from C4 news-like and English partitions~\citep{JMLR:v21:20-074}~\footnote{\url{huggingface.co/datasets/c4}}, and obtain around 61M synthetic triplets. In the domain adaptation setting, we follow TSDAE~\citep{wang-etal-2021-tsdae-using} to use unlabelled training set as in-domain sentences for AskUbuntu, CQADupStack, and Twitter. Since no training set is available for BIOSSES, we use PubMed subset in the Pile~\citep{gao2020pile} as in-domain sentences, and remove the sentences existing in the test set.

As for the QA training, we utilize public available QA data for training, since CommQA used in~\citep{abs-2108-08877} is not released. We choose datasets that are sampled from web sources, and have a sentence as both input and output. We also remove datasets that are closely related to downstream tasks, e.g., Stack Exchange, as well as the ones that are manually annotated, e.g., MS MARCO~\citep{bajaj2016ms}, for fair comparison. Finally, we obtain 4M QA pairs, including ELI5~\citep{fan-etal-2019-eli5}, GOOAQ~\citep{khashabi-etal-2021-gooaq-open}, and Yahoo~\citep{NIPS2015_250cf8b5}. 

We build GenSE upon the widely-used T5 Encoder-Decoder models~\citep{JMLR:v21:20-074}. For generator/discriminator training, we set the learning rate to 5e-5 and batch size to 256. The generator/discriminator is trained for 10 epochs with evaluation step set as 500, and $(accuracy-10 \times ppl)$ as validation metric for model selection. For data synthesis, we use nucleus sampling~\citep{holtzman2019curious} with $p=0.9$, and set the confidence threshold $\alpha=0.9$. For contrastive sentence representation learning, we set the learning rate to 5e-5 and batch size to 512 on NLI, and learning rate of 5e-5 with batch size of 1024 on the synthetic data. The temperature $\tau$ is set to 0.01 for QA training, and 0.05 for others.

\subsection{STS Results}
We compare GenSE with T5-Base to previous state-of-the-art sentence embedding methods, including encoder-based and encoder-decoder based models. All baselines are trained on the labeled NLI corpus. We also include large-scale models and models trained with additional large-scale semi-structured CommQA data for comparison. The results are reported in Table~\ref{tab:main_results}.

Overall, GenSE can greatly outperform previous state-of-the-art models. Specifically, GenSE achieves an average Spearman’s correlation of 84.78,  significantly outperforming Base-size sentence embedding models, and even surpassing methods with much larger model sizes, e.g., ST5-Enc-3B and ST5-Enc-3B-CommQA. GenSE attains new state-of-the-art base-model results on 6 out of 7 STS datasets, i.e., except the SICK-R tasks, demonstrating that our synthetic data can greatly improve sentence embedding quality with GenSE.  

We also report the performance of GenSE+, which achieves even higher performance with an average Spearman’s correlation of 85.19 by integrating additional QA data. 
The results suggest that our synthetic data is complementary to semi-structured data like question-answer pairs. But compared to semi-structured data which requires lots of efforts for data mining and cleaning, our GenSE only need  unlabelled sentences that are much easier to collect for various domains, exhibiting better practical applicability.

\begin{table*}
\centering
\small
\resizebox{\textwidth}{!}{

\begin{tabular}{lcccccccc}
\toprule
\multirow[t]{2}{*}{\textbf{Model}} & \multirow[t]{2}{*}{\textbf{Fine-tune Data}} & \multirow[t]{2}{*}{\textbf{AskU.}} & \multirow[t]{2}{*}{\textbf{CQADup.}} & \multicolumn{3}{c}{\textbf{TwitterP.}}        & \multirow[t]{2}{*}{\textbf{BIOSSES}} & \multirow[t]{2}{*}{\textbf{Avg.}} \\
\cmidrule{5-7}
 & & & & \textbf{TURL.} & \textbf{PIT} & \textbf{Avg.} & & \\
\midrule
\multicolumn{8}{c}{\textit{Direct transfer}}   \\
\midrule
SimCSE-BERT-Base    & NLI & 53.5 & 12.4 & 75.6 & 66.9 & 71.2 & 68.4 & 55.4 \\
SimCSE-RoBERTa-Base & NLI & 54.6 & 11.7 & 74.4 & 68.5 & 71.5 & 67.7 & 55.4 \\
ST5-Enc-Base        & NLI & 56.6 & 14.3 & 73.9 & 72.5 & 73.2 & 70.2 & 57.5 \\
ST5-EncDec-Base     & NLI & 56.1 & 13.7 & 73.3 & 75.0 & 74.1 & 71.4 & 57.9 \\
\midrule
GenSE    & Open-domain$\rightarrow$NLI & 58.2 & 15.3 & 76.3 & 75.9 & 76.1 & 73.1 & 59.8 \\
GenSE-QA & QA$\rightarrow$NLI          & 57.4 & 15.6 & 75.8 & 75.8 & 75.8 & 75.1 & 59.9 \\
GenSE+ & Open-domain$\rightarrow$QA$\rightarrow$NLI & \textbf{58.4} & \textbf{16.8} & \textbf{76.4} & \textbf{77.0} & \textbf{76.7} & \textbf{76.7} & \textbf{61.1} \\
\midrule
\multicolumn{8}{c}{\textit{Domain adaptation}}           \\
\midrule
SimCSE-BERT-Base    & In-domain  & 55.9 & 12.4 & 74.5 & 62.5 & 68.5 & 76.8 & 56.4 \\
SimCSE-BERT-Base    & In-domain$\rightarrow$NLI  & 56.2 & 13.1 & 75.5 & 67.3 & 71.4 & 76.9 & 57.8 \\
TSDAE-BERT-Base     & In-domain & 59.4 & 14.5 & 76.8 & 69.2 & 73.0 & 47.4 & 53.5 \\
TSDAE-BERT-Base     & In-domain$\rightarrow$NLI & 59.4 & 14.4 & 75.8 & 73.1 & 74.5 & 76.5 & 59.8 \\
\midrule
GenSE   & Open-domain$\rightarrow$In-domain & \textbf{60.3} & \textbf{16.2} & 75.0 & \textbf{77.3} & 76.2 & 77.8 & 61.3 \\
GenSE   & Open-domain$\rightarrow$In-domain$\rightarrow$NLI &  \textbf{60.3} & 16.0 & \textbf{76.7} & 76.7 & \textbf{76.7}  & \textbf{77.9} & \textbf{61.5} \\
\bottomrule
\end{tabular}
}

\caption{Performance on four transfer downstream tasks from various domains. Average precision (AP) is reported for AskUbuntu, CQADupStack, and Twitter. Spearman's correlation is reported for BIOSSES. The first block shows the result of out-of-the-box supervised sentence embeddings. The second block shows the result using different domain adaptation approaches. For ASkUbuntu, CQADupStack, and Twitter, the baseline results are from~\citep{wang-etal-2021-tsdae-using}. For BIOSSES, we obtain the baseline results using open-source repo~\protect\footnotemark.}
\label{tab:domain_results}
\end{table*}
\footnotetext{\url{github.com/princeton-nlp/SimCSE}\\\url{github.com/UKPLab/sentence-transformers}}

\subsection{Transfer Tasks}
\textbf{Direct transfer:} We first consider the direct transfer scenario where unlabeled in-domain data are not used for sentence embedding training, with results in the top of Table~\ref{tab:domain_results}. We can see that GenSE shows 1.9 \% improvements over strong ST5 baselines, and GenSE-QA works even slightly better. Combining QA and synthetic data, i.e., GenSE+, leads to a substantial improvement of 3.2\%, demonstrating the effectiveness of synthetic data and QA pairs.


\textbf{Domain adaptation:} We also evaluate the model for domain adaptation, where in-domain synthetic data is also included in contrastive learning. We can observe that the average performance is improve by 1.7\% over the best TSDAE-BERT-Base model through using in-domain data. The consistent and significant improvements over the four  tasks also demonstrate the great generalization ability across various domains of GenSE. Furthermore, compared to direct transfer, we can also achieve an improvement of 0.4\% even without labeled QA data, which convincingly demonstrates that GenSE can make full use of unlabeled sentences for better sentence embedding. 



\subsection{Ablation Studies}
In this section, we present ablation studies on each component in GenSE, including the prompt learning, model scale, amount of labeled or synthesized data, and synthesis strategy.

\begin{table}
\centering
\small
\begin{tabular}{lcc}
\toprule
\textbf{Model}  & \textbf{STS-B} & \textbf{Trasnfer} \\
\midrule
ST5-EncDec-Small & 86.0       & 55.9    \\
+Prompt          & 86.5       & 56.6    \\
\midrule
ST5-EncDec-Base  & 87.2       & 57.9    \\
+Prompt          & 87.7       & 58.2    \\
\bottomrule
\end{tabular}

\caption{Ablation study on prompt: performance on STS-B dev set, and average performance on four transfer tasks are shown.}
\label{tab:prompt_ablation}
\end{table}
\textbf{Prompt learning:} 
It has been shown that adding prompt can improve the encoder-based sentence embedding~\citep{abs-2201-04337}, but the impact of prompt in text-to-text sentence embedding remains unknown.  We compare the performance of models trained on NLI with or without prompt. As shown in Table~\ref{tab:prompt_ablation}, adding prompt consistently improve the performance for both STS-B and transfer tasks across different model sizes. Ablation experiments on T5-Small show that stacking a single decoder cannot further improve the performance. Yet, adding prompt can further boost the performance. We hypothesize that prompt helps to elicit knowledge contained in the decoder, leading to performance improvement.

\textbf{Model scale:} We investigate whether GenSE works across different model scales. Due to resource limitation, we conduct experiments on T5-Small. Specifically, we first train a unified data generator from T5-Small. Then we use the model for data synthesis, which results in 34M training pairs. Finally, we fine-tune a T5-Small sentence embedding model with synthetic and NLI data. Table~\ref{tab:model_scale} shows the result. GenSE outperforms ST5-Small by 1.6\% on average for STS tasks, and 1.1\% for transfer tasks. Yet, compared to T5-Base, the performance gain for T5-Small is less significant. It's reasonable since the small data augmenter is less expressive than base one, which produces less synthetic data with lower quality from same amount of unlabelled sentences. We also find that although adding decoder improves the sentence embedding for base and large size models, it brings no benefit for T5-small model. One possibility is that the T5-small decoder architecture is too shallow, which acts like a mean pooling in encoder-only model.
\begin{table}[ht]
\centering
\small
\begin{tabular}{lccc}
\toprule
\textbf{Model}  & \textbf{\# Params} & \textbf{STS} & \textbf{Transfer} \\
\midrule
ST5-Enc-Small    & 30M & 80.9 & 56.0  \\
ST5-EncDec-Small & 60M & 80.9 & 55.9  \\
\midrule
GenSE-Small      & 60M & \textbf{82.5} & \textbf{57.1}  \\             
\bottomrule
\end{tabular}

\caption{Performance of models using T5-Small: we report the average performance on STS and transfer benchmarks.}
\label{tab:model_scale}
\end{table}

\begin{table}
\centering
\small

\begin{tabular}{lccccc}
\toprule
\textbf{Data Amount} & 0\%  & 20\% & 40\% & 60\% & 100\% \\
\midrule
\textbf{STS-B}       & 87.7 & 88.7 & 88.8 & 88.7 & 88.9  \\
\textbf{Transfer}    & 58.2 & 59.3 & 59.5 & 59.7 & 59.8  \\
\bottomrule
\end{tabular}

\caption{Performance of GenSE with different amount of synthetic data. We report STS-B dev set and average transfer performance. 0\% refers to directly apply prompt-based contrastive learning on NLI.}
\label{tab:data_ablation}
\end{table}
\textbf{Synthetic data amount:} We also study how the amount of synthetic data influence the final performance. As shown in Table~\ref{tab:data_ablation}, by adding 20\% data, GenSE already achieves substantial improvement. GenSE achieves the best performance on STS-B dev set (88.9) with more data, which even outperforms fine-tuned cross-attention T5-Base (88.02). For the transfer tasks, the average performance continues to improve as more data are used. The experiments validate the idea of using open-domain sentences to improve model generalization ability.

\textbf{Supervised data amount:} We investigate whether synthetic data can help reduce the annotation burden. 
We train another GenSE model with 50\% randomly-sampled NLI data, with results in Table~\ref{tab:sup_data_ablation}. Together with  Table~\ref{tab:prompt_ablation} we can find that the performance of baselines degrades significantly, i.e., 0.6\% on STS-B and 1.4\% on transfer tasks. However, using synthetic data can greatly alleviate the performance drop, and even outperforms baseline models trained on full NLI data.

\begin{table}
\centering
\small
\begin{tabular}{lcc}
\toprule
\textbf{Model}  & \textbf{STS-B} & \textbf{Trasnfer} \\
\midrule
ST5-EncDec-Base + Prompt  & 87.1 & 56.8 \\
GenSE                     & 88.6 & 59.1 \\
\bottomrule
\end{tabular}

\caption{Performance of models trained with 50\% of supervised NLI data. Performance on STS-B dev set and transfer tasks are shown.}
\label{tab:sup_data_ablation}
\end{table}



\begin{table}
\centering
\small
\begin{tabular}{lcc}
\toprule
\textbf{Model}  & \textbf{STS-B} & \textbf{Transfer} \\
\midrule 
ST5-EncDec-Base + Prompt      & 87.7 & 58.2 \\
\midrule
Retrieval                     & 88.3 & 58.9 \\
Back Translation                 & 88.0 & 59.6 \\
\midrule
GenSE                         & 88.9 &  61.5  \\
w/o synthetic negatives       & 88.6 & 59.5 \\
\bottomrule
\end{tabular}

\caption{Comparison of different synthesis strategies on STS-B dev set and transfer tasks}
\label{tab:synthesis_strategy}
\end{table}

\begin{table*}[ht]
\centering
\small
\resizebox{\textwidth}{!}{
\begin{tabular}{l|l|l}
\toprule
\textbf{Method}       & \textbf{Entailment}    & \textbf{Contradiction}     \\
\midrule
& \multicolumn{2}{l}{\textit{\textbf{Input:} A young man getting ready to release a red kite.}}  \\
\midrule
\multirow{2}{*}{DINO} & A young man releasing a red kite. & A man getting ready to release a red kite. \\
& A red kite releasing a red kite. & It was a big deal to him and he didn't know how he would explain it to his parents. \\
\midrule
\multirow{2}{*}{Retrieval} & A man getting ready to fly a kite.                & N/A \\
                           & A man in a yard getting ready to play with a kite & N/A \\
\midrule
\multirow{2}{*}{BackTrans} & A young man prepares to release a red kite.       & N/A \\
                           & A young man is about to release a red kite.       & N/A \\
\midrule
\multirow{2}{*}{GenSE}     & The man is prepared to fly the kite.  & A man is playing basketball. \\
                           & A man is planning to fly a kite. & The woman is flying a kite. \\
\midrule
& \multicolumn{2}{l}{\textit{\textbf{Input:} One of the hotel's rooms}}  \\
\midrule
\multirow{2}{*}{DINO} & The hotel room          & I have no idea what that is.  \\
                      & One of the hotel rooms  & The other one is on fire \\
\midrule
\multirow{2}{*}{Retrieval} & One of the 300 modular hotel rooms         & N/A  \\
                           & The grand hotel birmingham one of the hotel rooms & N/A \\
\midrule
\multirow{2}{*}{BackTrans} & One of the hotel rooms          & N/A  \\
                           & One of the rooms of the hotel   & N/A  \\
\midrule
\multirow{2}{*}{GenSE} & A room inside a hotel. & There is no room at the hotel.  \\
                           & A hotel room.          & It's not the hotel's room.      \\
\bottomrule
\end{tabular}
}

\caption{Comparison of different data synthesis approaches. For retrieval-based method, only entailment pairs can be obtained. We bold the correct samples through human judgement.}
\label{tab:full_gen_samples}
\end{table*}
\textbf{Data synthesis strategy:} 
We demonstrate the superiority of our data synthesis strategy by comparing to previous methods that utilize synthetic data for sentence embedding under the same semi-supervised setting, i.e.,back-translation~\citep{wieting-gimpel-2018-paranmt,zhang-etal-2021-bootstrapped} and retrieval~\citep{thakur-etal-2021-augmented}. The results are in Table~\ref{tab:synthesis_strategy}. We can see that both retrieval-based method and back-translation can improve the performance, while our generator/discriminator based method can further significantly boost the performance on both tasks. Further experiments for GenSE without hard negatives demonstrates that synthesizing negative pairs plays an important role for the performance improvement, as it provides more information for contrastive learning. 


\subsection{Analysis}
\textbf{Uniformity and alignment:} We investigate the uniformity and alignment of GenSE, which capture the quality of produced sentence embedding:
\begin{equation}
    \mathcal{L}_{align} = - \mathop{\mathbb{E}}_{s,s^+ \sim p_{pos}}\Arrowvert f(s)-f(s^+)\Arrowvert,
\end{equation}
\begin{equation}
    \mathcal{L}_{uniform} = \log \mathop{\mathbb{E}}_{s,u \overset{\mathrm{iid}}{\sim} p_{data}} e^{-2\Arrowvert f(s)-f(u)\Arrowvert},
\end{equation}
where $p_{data}$ is the distribution of positive pairs, and $p_{pos}$ refers to all entailment pairs. Smaller $\mathcal{L}_{align}$ indicates shorter distance between sentence embeddings of positive pairs, and smaller $\mathcal{L}_{uniform}$ means that the embedding space is more uniform.

\begin{figure}[t]
    \centering
    \includegraphics[scale=0.50]{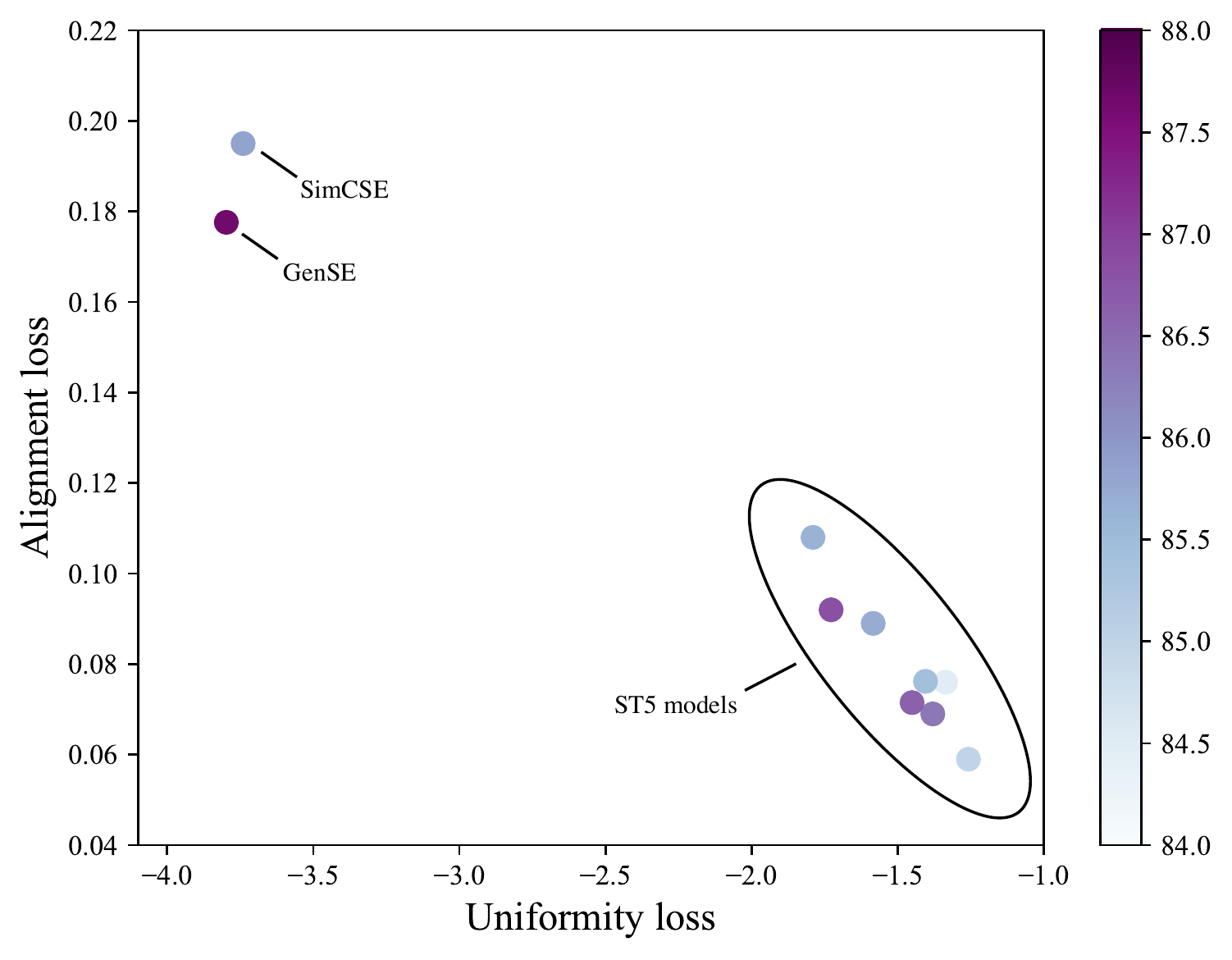}
    \vspace{-1mm}
    \caption{Alignment and uniformity losses plot: The color of dots refer to models' performance on STS-B test split. For ST5 models, we reports the result from the original paper.~\citep{abs-2108-08877}} 
    \vspace{-3mm}
    \label{fig:align_uniform}
\end{figure}

We follow the setting in ~\citep{abs-2108-08877} to measure $\mathcal{L}_{uniform}$ on the whole STS-B test set, and $\mathcal{L}_{align}$ on sentence pairs in STS-B test set with correlation scores higher than 4. 
As shown in Figure~\ref{fig:align_uniform}, GenSE shows better alignment and uniformity than the SimCSE model based on RoBERTa-Base. Compared to the supervised ST5 baseline models, GenSE achieves much lower uniformity loss but larger  alignment loss. We conjecture that GenSE obtains lower uniformity by learning from more synthesized unlabeled data, which might bring some noisy pairs and result in large alignment loss. 

\textbf{Quality of synthetic data:} Several methods have been implemented to synthesize the sentence pairs, including generation-based DINO~\citep{schick-schutze-2021-generating}, retrieval-based approach~\citep{thakur-etal-2021-augmented}, and back-translation, with samples shown in Table~\ref{tab:full_gen_samples}. For each input sentences, we give two samples using over-generation. We run the experiment on the image caption data CC12M~\citep{changpinyo2021conceptual}. For DINO, we use the official repo~\footnote{\url{https://github.com/timoschick/dino}}. For retrieval, we follow~\citep{thakur-etal-2021-augmented} to use BM25~\citep{Amati2009} to produce possible pairs, and use a cross-encoder trained on NLI to further label the pair. For back-translation, we use google translator to produce English-French and English-German pairs.

For entailment pair generation, unsupervised DINO~\citep{schick-schutze-2021-generating} can generate some meaningful pairs with a large-scale generative model. It also produces many incorrect pairs, and cannot be filtered out since no supervision signal is available. BackTrans, Retrieval and GenSE can all produce entailments efficiently. However, BackTrans usually produces entailment pairs with very high lexical overlap, which fail to give high-quality supervision signals for contrastive representation learning. Retrieval can produce noisy pairs even after filtering due to the limited size of the banking corpus.

For contradiction pair generation, DINO hardly generate correct or related sentences, which cannot serve as hard negatives in sentence representation learning. Although retrieval and back-translation can produce entailment pairs, they cannot generate contradiction pairs. In contrast, GenSE can produce both entailment and contradiction pairs, which are important for contrastive sentence representation learning as demonstrated in Table~\ref{tab:synthesis_strategy}.

\section{Related Work}
Prior approaches for sentence embedding include two main categories: (1) supervised learning with labeled sentences, and (2) unsupervised sentence embedding with unlabeled sentences, while a few early approaches leverage on both of them.

Supervised sentence representation learning relies on human-annotated sentence pairs, e.g., NLI data. Early works learn sentence embedding through fine-tuning the model on NLI with classification objectives~\citep{conneau-etal-2017-supervised,cer2018universal}.  Recent works find that contrastive objectives can help learn better sentence representation~\citep{gao-etal-2021-simcse,carlsson2020semantic}. There have been several works exploring the effect of additional supervised training pairs on sentence representation learning~\citep{abs-2108-08877}. ST5~\citep{abs-2108-08877} utilizes question-answer pairs for pre-training before fine-tuning the model on the NLI corpus, leading to better generalization performance.

Many approaches attempt to develop unsupervised objectives for sentence embedding.
Early works train the model to predict surrounding sentences~\citep{kiros2015skip, hill-etal-2016-learning,logeswaran2018efficient}. Recent works start to adopt contrastive learning through maximizing the similarity between different view of the same sentences~\citep{zhang-2020-unsupervised,carlsson2021semantic, giorgi-etal-2021-declutr,gao-etal-2021-simcse}. Recently,~\citet{abs-2201-04337} utilizes the prompt to extract embeddings from encoder models, which inspires the prompt-based contrastive objective in our GenSE. Despite the promising results from unsupervised approaches, there's still a large performance gap between unsupervised and supervised approaches.

In this work, we aim to combine the supervised and unsupervised approach. Similar to our motivation, USE~\citep{cer2018universal} uses the SkipThought-like~\citep{kiros2015skip} loss for unlabeled sentences, and a classification loss for NLI. However, the performance is unsatisfactory. Recent works mainly focus on using unlabeled data for domain adaptation.  \citet{thakur-etal-2021-augmented} first adopts sampling strategies, e.g. BM25~\citep{Amati2009} and semantic search, to obtain weakly-labelled pairs, and then uses cross-encoders trained on NLI for re-labelling. \citet{wang-etal-2021-tsdae-using} proposes an auto-encoder loss for unsupervised domain adaptation of supervised sentence encoder. Different from these approaches, we utilize data synthesis model~\citep{shorten2019survey,9020132,feng-etal-2021-survey,gao2021auto,gao2022genre} to convert large-scale unlabeled sentences into sentence pairs towards better sentence embeddings.

\section{Conclusion and Future Work}
In this work, we propose a novel semi-supervised framework, GenSE, for sentence representation learning. We first train a unified model for generation and discrimination, which can effectively obtain high-quality synthetic positive and negative sentence pairs from open-domain unlabelled corpus. Afterwards, we train a prompt-based text-to-text sentence embedding model with contrastive learning on both synthesized and labeled NLI data. Extensive results on STS and transfer tasks validate that GenSE can achieve significantly better performance than current state-of-the-arts and exhibit better generalization ability. 
Future work includes better synthesizing strategies to generate better sentence pairs and advanced designs of semi-supervised sentence representation learning frameworks based on more diverse open-domain data. 
\section{Limitations}
Firstly, our generator/discriminator is only trained on NLI data, which makes the generator/discriminator less expressive. As demonstrated in ST5 and GenSE, using semi-structured data, e.g., QA pairs, in contrastive sentence representation learning leads to a significant performance improvement, especially on transfer tasks. We hypothesize that QA pairs contain rich information, and capture very different semantic relationships compared to NLI pairs, which could lead to much stronger generalization ability. Therefore, we plan to include semi-structured data into generator/discriminator training as future work to produce more diverse synthetic pairs.


Secondly, due to insufficient GPU resources, we are unable to scale our model to T5-Large, or using more unlabelled data for training. Therefore, we cannot fully evaluate the potential of synthetic data. In addition, we cannot include large-scale QA data in training, which will lead to a more universal sentence embedding.

Thirdly, multi-stage training on the synthetic data leads to higher computational cost. Training on ~61M synthetic triplets takes around 88 GPU hours. As shown in the ablation study on the amount of synthetic data , GenSE continues to improve with more data. To achieve a better trade-off between performance and cost, we leave representative data selection and efficient sentence embedding training as future research directions.
\section*{Acknowledgements}
We would like to thank all the reviewers for their constructive comments. This work is supported by Science and Engineering Research Council, Agency of Science, Technology and Research (A*STAR), Singapore, through the National Robotics Program under Human-Robot Interaction Phase 1 (Grant No. 192 25 00054); Human Robot Collaborative AI under its AME Programmatic Funding Scheme (Project No. A18A2b0046); This work is partially supported by the Internal Project Fund from Shenzhen Research Institute of Big Data under Grant T00120220002; This work is also supported by the National Natural Science Foundation of China (Grant No: 62106222).

\bibliography{anthology,custom}
\bibliographystyle{acl_natbib}



\end{document}